\documentclass[letterpaper, 10 pt, conference]{ieeeconf}
\IEEEoverridecommandlockouts
\usepackage{graphics} 
\usepackage{epsfig} 
\usepackage{mathptmx} 
\usepackage{amsmath} 
\usepackage{amssymb}  
\usepackage{booktabs}
\usepackage{multirow}
\usepackage{makecell}
\usepackage{siunitx}

\title{\LARGE \bf
GRATE: a Graph transformer-based deep Reinforcement learning Approach for Time-efficient autonomous robot Exploration}

\author{Haozhan Ni$^{1}$, Jingsong Liang$^{1}$, Chenyu He$^{2}$, Yuhong Cao$^{\SI{1}{\dagger}}$, Guillaume Sartoretti $^{1}$
\thanks{$^{\dagger}$ Corresponding author, to whom correspondence should be addressed.}
\thanks{$^{1}$ Department of Mechanical Engineering, National University of Singapore, Singapore. {\tt\small nihaozhan@u.nus.edu, \{jsliang, caoyuhong, mpegas\}@nus.edu.sg}}
\thanks{$^{2}$ School of Automotive Studies, Tongji University, Shanghai, China. {\tt\small 2152393@tongji.edu.cn}}
}

\begin{document}

\maketitle
\thispagestyle{empty}
\pagestyle{empty}

\begin{abstract}

Autonomous robot exploration (ARE) is the process of a robot autonomously navigating and mapping an unknown environment. Recent Reinforcement Learning (RL)-based approaches typically formulate ARE as a sequential decision-making problem defined on a collision-free informative graph. However, these methods often demonstrate limited reasoning ability over graph-structured data. Moreover, due to the insufficient consideration of robot motion, the resulting RL policies are generally optimized to minimize travel distance, while neglecting time efficiency. To overcome these limitations, we propose GRATE, a Deep Reinforcement Learning (DRL)-based approach that leverages a Graph Transformer to effectively capture both local structure patterns and global contextual dependencies of the informative graph, thereby enhancing the model’s reasoning capability across the entire environment. In addition, we deploy a Kalman filter to smooth the waypoint outputs, ensuring that the resulting path is kinodynamically feasible for the robot to follow. Experimental results demonstrate that our method exhibits better exploration efficiency (up to 21.5\% in distance and 21.3\% in time to complete exploration) than state-of-the-art conventional and learning-based baselines in various simulation benchmarks. We also validate our planner in real-world scenarios.

\end{abstract}

\section{INTRODUCTION}

Autonomous Robot Exploration (ARE) involves a robot actively perceiving its surroundings, making informed decisions under uncertainty, and continuously updating its map in real time. Typically, the robot is equipped with sensors (e.g., camera(s) or LiDAR) to map its surroundings. In this work, we focus on autonomous exploration using an omnidirectional 3D LiDAR, operating within a 2D action space (i.e., for a ground robot). The objective of ARE is to explore the whole environment while minimizing travel distance and exploration time.

The main challenge in ARE is to plan non-myopic paths that effectively balance exploiting known regions with exploring unknown areas. Conventional frontier-based methods~\cite{yamauchi1997frontier, gonzalez2002navigation, osswald2016speeding, cieslewski2017rapid} typically guide the robot greedily toward nearby frontiers, gradually uncovering unexplored areas through an information gain function. However, such myopic frontier selection provides no guarantee of long-term optimality~\cite{cao2023ariadne}. To mitigate this, advanced conventional sampling-based methods~\cite{bircher2016NBVP, cao2021tare, cao2023representation, huang2023fael, long2024hphs} incorporate hierarchical strategies that combine local and global planning. Nevertheless, these methods still share a fundamental limitation: they optimize trajectories based on the partial robot belief (map). As the map is incrementally refined during exploration, decisions made from this incomplete view may ultimately be suboptimal~\cite{cao2024deep}.

\begin{figure}
    \centering
    \includegraphics[width=1\linewidth]{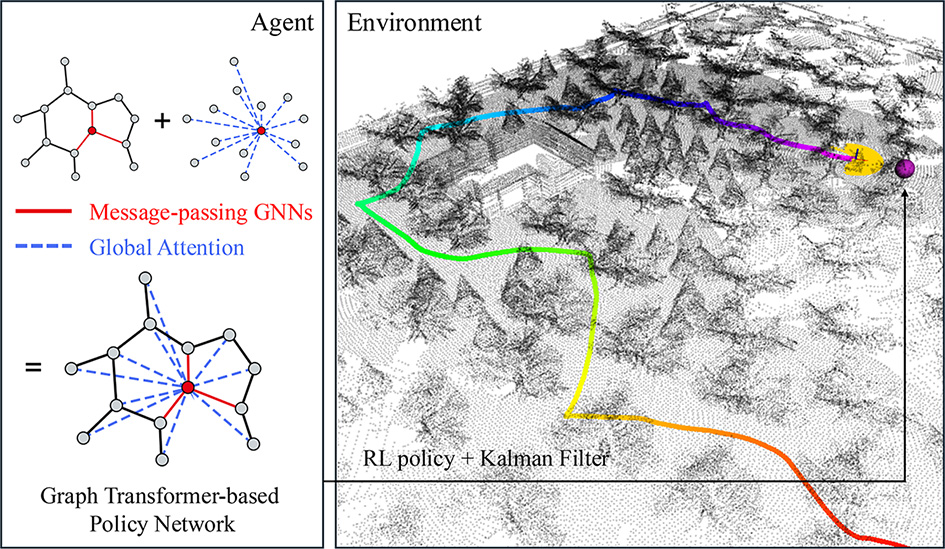}
    \vspace{-0.75cm}
    \caption{\textbf{Illustration of GRATE, our Graph Transformer-based DRL approach for autonomous robot exploration.} Left: Our proposed Graph Transformer-based policy network, integrating message-passing GNNs with attention to capture local and global dependencies. Right: Visualization results of GRATE in a large-scale simulated forest environment from the exploration benchmark provided by \cite{cao2022autonomous}. The next waypoint (purple ball) is determined by integrating the probability outputs of both our RL policy and Kalman filter.}
    \label{fig_1}
    \vspace{-0.4cm}
\end{figure}

To address this issue, current learning-based approaches~\cite{cao2023ariadne, cao2024deep, liang2024hdplanner, chen2020autonomous} formulate the ARE problem as a sequential decision-making task defined on a collision-free informative graph, leveraging deep reinforcement learning (DRL) to implicitly reason about unexplored regions from the topological structure of the explored area. The effectiveness of these approaches largely depends on designing a framework with strong reasoning capability. For instance, inspired by Graph Attention Networks (GAT)~\cite{velickovic2018gan}, current state-of-the-art (SOTA) learning-based approaches~\cite{cao2023ariadne, cao2024deep, liang2024hdplanner} employs structurally masked attention guided by graph adjacency, which restricts each node’s receptive field and requires stacking multiple layers for global propagation. While effective on small-scale graphs, this design suffers from over-squashing ~\cite{alon2021bottleneck, topping2022understanding} in large-scale graphs. Over-squashing occurs when information from many distant nodes is overly compressed into a fixed-size node feature, hindering global structure representation and under-representing faraway nodes. Meanwhile, pure global attention in Transformers~\cite{vaswani2017attention} overlooks the structural information encoded by graph adjacency (indicating which nodes are directly connected). Graph Transformers~\cite{wu2021representing, chen2022structure, rampavsek2022recipe, wu2023sgformer} address these limitations by integrating local message-passing graph neural networks (GNNs) ~\cite{dwivedi2023benchmarking} with Transformers, capturing both local structure patterns and global contextual dependencies. Inspired by this, we propose a Graph Transformer-based DRL approach for ARE, enhancing the model’s ability to reason over both explored and unexplored regions.

Another limitation of current learning-based approaches \cite{cao2023ariadne, cao2024deep, liang2024hdplanner, chen2020autonomous} is their insufficient consideration of robot kinematics. The core issue is that training typically accounts only for the robot’s position transitions while neglecting key physical attributes of robot motion, such as orientation, velocity, and acceleration. For example, these approaches model the robot as a node in the collision-free informative graph. At each decision step, the robot selects one of its neighboring nodes as the next waypoint and is assumed to transition to it instantaneously. As a result of these motion discontinuities, Reinforcement Learning (RL) models trained under such assumptions are generally optimized to minimize travel distance while overlooking time efficiency. Consequently, the resulting policies often produce zig-zag or conflicting waypoints. Although these behaviors may be spatially optimal in the training environment, they often lead to poor temporal efficiency in high-fidelity simulation environments and real-world applications. The most direct way to address this issue is to incorporate a local motion planner and use it to inform or train the policy. However, introducing robot motion constraints, a highly complex and uncertain system, substantially reduces training stability and convergence speed. To address this issue, we apply a Kalman filter to smooth the waypoint outputs of the RL policy. Historically selected waypoints serve as observations to estimate the robot’s velocity and position, guiding our planner to generate a kinodynamically feasible path. Here, the policy is trained in the same way as in existing works \cite{cao2023ariadne, cao2024deep}. We primarily post-process its outputs to improve time efficiency.

Leveraging these new ideas and developments, we propose GRATE, a Graph Transformer-based DRL approach for autonomous robot exploration. We evaluate GRATE against SOTA conventional (TARE~\cite{cao2021tare, cao2023representation}, FAEL~\cite{huang2023fael}, HPHS~\cite{long2024hphs}) and learning-based (ARiADNE~\cite{cao2023ariadne, cao2024deep}) approaches in various high-fidelity Gazebo simulation benchmarks. Experimental results demonstrate that GRATE significantly outperforms these baselines in terms of exploration efficiency (travel distance and exploration time). We also experimentally validate our planner on a ground robot in real-world scenarios, highlighting its sim-to-real applicability.

\section{Related Works}

\subsection{Conventional Exploration Planners}
Conventional exploration planners can be broadly categorized into frontier-based and sampling-based methods. Frontier-based approaches identify regions at the boundary between explored and unexplored areas, which are commonly used to quantify information gain and guide the exploration process. For example, conventional frontier-based works~\cite{yamauchi1997frontier, gonzalez2002navigation, osswald2016speeding, cieslewski2017rapid} select the frontier to visit through a information gain function, quantifying the amount of information expected to be collected at that frontier. More recent sampling-based works~\cite{bircher2016NBVP, cao2021tare, cao2023representation, huang2023fael, long2024hphs} typically adopt carefully designed sampling strategies to balance planning quality and computational efficiency. In particular, TARE~\cite{cao2021tare, cao2023representation} integrates a Traveling Salesman Problem (TSP)-based global planner with a sampling-based local planner to generate exploration paths that are optimized for both efficiency and kinodynamic feasibility. FAEL \cite{huang2023fael} optimizes runtime through fast environmental preprocessing and heuristic path optimization, enabling high-frequency planning and surpassing SOTA methods in exploration efficiency. HPHS \cite{long2024hphs} is a fast exploration framework that integrates hybrid frontier sampling with hierarchical planning, ensuring scalable sampling efficiency while reducing planning complexity and mitigating region remnants.

\vspace{-0.2cm}
\subsection{Learning-based Exploration Planners}
Current DRL-based approaches \cite{cao2023ariadne, cao2024deep, liang2024hdplanner, chen2020autonomous, zhu2018deep, chen2019self} typically formulate the ARE problem as a sequential decision-making task, leveraging various neural networks to implicitly reason over the entire environment and learn an adaptive policy. ~\cite{zhu2018deep, chen2019self} applied CNNs to encode exploration maps and determine the robot’s action through DRL. Chen et al.~\cite{chen2020autonomous} utilize GNNs in conjunction with DRL to predict a robot’s optimal next waypoint in belief space. Notably, by leveraging the attention mechanism, recent DRL-based approaches~\cite{cao2023ariadne, cao2024deep, liang2024hdplanner} have demonstrated superior performance compared to both conventional and learning-based SOTA exploration planners. More recent works \cite{wang2025cogniplan, harutyunyan2025mapexrl} integrate explicit map prediction with implicit policy learning. \cite{wang2025cogniplan} first explicitly predicts the map layouts by a generative inpainting model and then implicitly reasons about and plans paths through a graph-attention-based DRL planner. Similarly, \cite{harutyunyan2025mapexrl} generates global map predictions from the exploration map via a CNN-based prediction module while simultaneously training a frontier point selection policy to enable longer-horizon decision making.

\vspace{-0.2cm}
\subsection{Graph Transformers}
Transformers~\cite{vaswani2017attention} can be viewed as a specific type of GNNs, where self-attention is applied over all nodes in a fully connected graph~\cite{yuan2025survey}. Since Transformers do not account for the underlying graph structure, Graph Transformers are designed to integrate structural priors directly into the Transformer architecture~\cite{yuan2025survey, ying2021transformers}. Based on the approaches of incorporating this structural prior, we can categorize existing Graph Transformers into~\cite{yuan2025survey}: (1)  structural positional encoding~\cite{ying2021transformers, dwivedi2020generalization}, (2) structure-aware attention mechanisms~\cite{ying2021transformers, dwivedi2020generalization}, (3) hybrid architectures that integrate GNNs into Transformers~\cite{wu2021representing, chen2022structure, rampavsek2022recipe, wu2023sgformer}. Notably, GraphGPS~\cite{rampavsek2022recipe} achieves highly competitive performance across 16 benchmarks by combining GNNs with Transformers. Furthermore, SGFormer~\cite{wu2023sgformer} critically demonstrates that coupling one-layer global attention with GNNs, without any positional encoding, can still significantly outperform conventional GNNs (i.e., GCN, GAT). Overall, Graph Transformers outperform traditional GNNs and Transformers because they retain the graph’s local structural information while capturing long-term dependencies through global attention, enabling stronger reasoning and representation capabilities.

\section{Problem Formulation}

The environment is represented as a bounded discrete 2D grid map, capturing the spatial layout of the exploration area. The ground truth map 
\( G \in \{0,1\}^{W \times H} \) indicates the occupancy of each cell, where \( G_{i,j}=1 \) corresponds to an obstacle and \( G_{i,j}=0 \) to free space. 
The robot maintains a belief map \( B_t \in \{0,0.5,1\}^{W \times H} \) at time step \( t \), which is updated based on observations. 
Each cell \( B_{t,i,j} \) encodes the occupancy probability: \(0.5\) for unknown (\(B_u\)), \(0\) for free (\(B_f\)), and \(1\) for occupied (\(B_o\)). The exploration is considered complete when \( B_o \) is closed with respect to \( B_f \).

The ARE problem can be modeled as a sequential decision-making problem. In this task, a completed exploration trajectory is represented as a finite sequence of robot states \( \psi = [s_1, \dots, s_t, s_{t+1}, \dots, s_T] \), with \(s_t\) indicating the robot state at time step \(t\). During exploration, the robot only has access to observations \(o_t\). At each decision step, the robot selects an action \(a_t\) according to the policy \( \pi \), where \( a_t  \sim \pi(\cdot | o_t) \). Upon transitioning to the next state \(s_{t+1}\), the agent receives a reward \(r_t\). The objective of ARE is to find an optimal policy \(\pi^*\) that maximizes the expected cumulative reward \( E\left( \sum_{t=1}^{T} \gamma^{t-1} r_t \right) \), while simultaneously minimizing the total distance traveled along the exploration trajectory \(\psi\).

\section{Methodology}

Our method builds upon \cite{cao2023ariadne, cao2024deep}, which represents the robot’s belief \(B_t\) as a collision-free informative graph. To enhance the model’s reasoning capability over a graph, we propose GRATE, a DRL framework for autonomous robot exploration. GRATE leverages Graph Transformer, a hybrid architecture that combines message-passing GNNs with multi-head self-attention to capture both local and global dependencies. The local message-passing module encodes neighboring information and preserves spatial relationships, while global attention incorporates broader graph context, enabling more informed decision-making and improving exploration efficiency over complex environments.

We further integrate a Kalman filter to account for robot motion, smoothing the sequence of selected waypoints and generating kinodynamically feasible paths in both high-fidelity simulations and real-world scenarios. Specifically, previously selected waypoints are treated as observations, allowing the Kalman filter to predict a probabilistic distribution of the robot position. This prediction is then integrated with the output of the RL policy to determine the final next waypoint, effectively balancing the policy’s exploration guidance with the robot’s motion feasibility. By doing so, our approach mitigates zig-zag or conflicting waypoints, resulting in paths that are smoother and more time-efficient for real-world deployment.

\begin{figure*}
    \centering
    \includegraphics[width=0.9\linewidth]{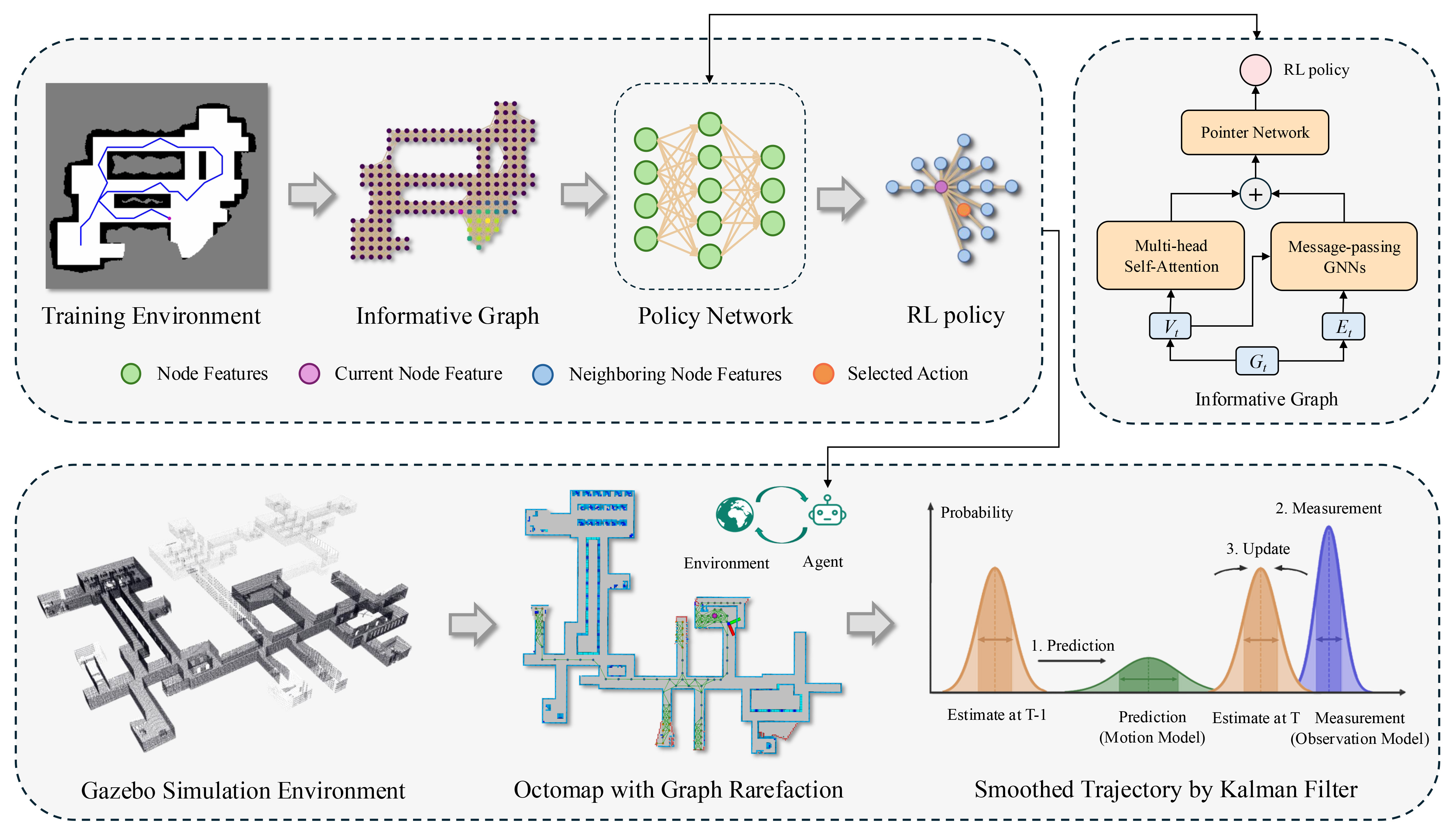}
    \vspace{-0.5cm}
    \caption{\textbf{Our proposed time-efficient Graph Transformer-based DRL approach for autonomous robot exploration.} In the training environment~\cite{chen2019self}, current robot's belief is extracted as a collision-free informative graph. This graph is then encoded by our proposed Graph Transformer-based policy network and decoded by Pointer Network~\cite{vinyals2015pointer} to generate the RL policy. In the large-scale high-fidelity Gazebo Simulation Environments~\cite{cao2022autonomous}, current robot's belief (point cloud data) is transformed into Octomap~\cite{hornung2013octomap}, which is then extracted as a sparse informative graph~\cite{cao2024deep} to extend the use of RL policy from small-scale to large-scale environments. We finally apply a Kalman filter to smooth the trajectory output of the RL policy.}
    \label{fig_2}
    \vspace{-0.5cm}
\end{figure*}

\subsection{Graph Representation of Robot Belief}
At each time step \(t\), the robot's current belief \(B_t\) is extracted as a collision-free informative graph \(G_t = (V_t, E_t)\), where the node set \(V_t = \{v_1, v_2, \dots, v_n\} \in B_f\) consists of candidate waypoints. Each node \(v_i = [x_i, y_i, u_i, g_i]\) encodes: (1) coordinates \((x_i, y_i)\); (2) utility \(u_i\), which quantifies the number of observable frontiers from \(v_i\); and (3) a guidepost \(g_i\), a binary indicator specifying whether \(v_i\) has been previously visited. The edge set \(E_t\) is represented by a symmetric (typically sparse) adjacency matrix \(A_t = [a_{i,j}]\), where \(a_{i,j} = 0\) indicates a collision-free connection between nodes \(v_i\) and \(v_j\), and \(a_{i,j} = 1\) otherwise. To construct collision-free edges, each waypoint is connected to its \(k\)-nearest neighbors (\(k = 25\) in practice) using straight-line paths, and edges intersecting occupied (\(B_o\)) or unknown (\(B_u\)) cells are removed. Candidate waypoints in \(V_t\) are uniformly sampled to cover all free space \(B_f\). Before feeding the graph into the Graph Transformer-based policy network, both the node coordinates and utility values are normalized to the range \([0,1]\).

\subsection{Policy Network}
\noindent \textbf{Encoder} The encoder is responsible for extracting features from the input informative graph, which consists of nodes \( V_t \in \mathbb{R}^{N \times 4}\) and adjacency matrix \( A_t \in \mathbb{R}^{N \times N}\), where \( N \) is the number of nodes. Before feeding data into the encoder, we first project the node inputs \(V_t \in \mathbb{R}^{N \times 4}\) into \( d \)-dimensional node features \(X_t \in \mathbb{R}^{N \times d}\) using a linear feed-forward neural network. The encoder consists of multiple (3 in practice) encoder layers, with each encoder layer combining m-layer global multi-head self-attention and n-layer local message-passing GNN blocks (in practice, \(m=1, n=2\)) to capture global and local dependencies.
\begin{align}
    \mathbf{X}^{\ell+1}_t &= \mathrm{Encoder}^{\ell}(\mathbf{X}^{\ell}_t, \mathbf{A}_t) \notag \\
    \text{computed as} \quad
    \mathbf{X}^{\ell+1}_A &= \mathrm{MLP}^{\ell}(\mathrm{Attn}^{\ell}(\mathbf{X}^{\ell}_t)) \\
    \mathbf{X}^{\ell+1}_G &= \mathrm{MLP}^{\ell}(\mathrm{GNN}^{\ell}(\mathbf{X}^{\ell}_t, \mathbf{A}_t)) \notag \\
    \mathbf{X}^{\ell+1}_t &= (1-\alpha) \mathbf{X}^{\ell+m}_A + \alpha \mathbf{X}^{\ell+n}_G \notag,
\end{align}
where \(\mathrm{Attn}^{\ell}\) and \(\mathrm{GNN}^{\ell}\) are instances of multi-head self-attention and message-passing GNN blocks at the \(\ell\)-th layer with their corresponding learnable parameters; \(\mathrm{MLP}^{\ell}\) is a 2-layer feed-forward neural network; \(\alpha\) is a weight hyper-parameter (in practice, \(\alpha = 0.8\)).

\noindent \textbf{Multi-head Self-Attention} In contrast to~\cite{cao2023ariadne, cao2024deep,liang2024hdplanner}, which utilized a masked attention mechanism to restrict each node’s receptive field to its local neighborhood and required stacking multiple layers to propagate information globally, we employ an all-pair multi-head self-attention layer in vanilla Transformer~\cite{vaswani2017attention}. This design enables direct modeling of global contextual dependencies by computing pairwise interactions between arbitrary nodes. An attention layer receives a query vector \(h_i^q\) and a key-and-value vector \(h_i^{k,v}\) as input. Its output \(h_i'\) is obtained by taking a weighted combination of the value vectors, where the weights reflect the similarity between the query and key.
\begin{alignat}{3}
    q_i &= W^Q h^q_i,        &\quad k_i &= W^K h^{k,v}_i,     &\quad v_i &= W^V h^{k,v}_i, \notag \\
    u_{ij} &= \frac{q_i k_j^T}{\sqrt{d}}, \quad
    & w_{ij} &= \frac{e^{u_{i,j}}}{\sum_{j=1}^n e^{u_{i,j}}}, \quad
    & h'_i &= \sum_{j=1}^n w_{ij} v_j,
\end{alignat}
where \( W^Q, W^K, W^V \in \mathbb{R}^{d \times d} \) are all learnable matrices.

\noindent \textbf{Message-passing GNNs} GNNs~\cite{dwivedi2023benchmarking} generally operate through a recursive message-passing mechanism, in which each node updates its feature vector by aggregating information from its neighbors. After \(k\) iterations of aggregation, a node’s resulting feature vector encodes the structural information from its \(k\)-hop neighborhood~\cite{xu2018powerful}. In this work, we adopted ResGatedGCN~\cite{bresson2017residual}, a residual graph convolutional neural network incorporating the gated mechanism from LSTM~\cite{hochreiter1997long}:
\begin{gather}
    h^{\ell+1}_i = W_1 h^{\ell}_i + \sum_{j \in \mathcal{N}(i)} \eta_{i,j} \odot W_2 h^{\ell}_j, \notag \\
    \eta_{i,j} = \sigma \left(W_3 h^{\ell}_i + W_4 h^{\ell}_j \right),
\end{gather}
where \(h^{\ell}_i\) denote the feature vector at layer \(\ell\) associated with node \(i\). The updated features \(h^{\ell+1}_i\) at the next layer \(\ell+1\) are obtained by aggregating the feature vectors for all nodes \(j\) in the neighborhood of node \(i\) (defined by the adjacency matrix \(A_t\)). \( W_1, W_2, W_3, W_4 \in \mathbb{R}^{d \times d} \) are all learnable matrices; \(\odot\) is the Hadamard product. \(\sigma\) denote the sigmoid function.

\noindent \textbf{Decoder} The decoder layer applies a single-head attention mechanism, inspired by the Pointer Network~\cite{vinyals2015pointer}, to emphasize informative neighboring nodes during decision making. In this formulation, the enhanced feature of the current node acts as the query \(q_i\), while features of its neighbors serve as both keys \(k_j\). The attention module computes similarity scores between the query and each key, which are normalized into a probability distribution over neighbors. These attention weights \(w_{ij}\) reflect the likelihood of selecting each neighbor, guiding the robot’s next waypoint.
\begin{alignat}{3}
    \quad u_{ij} & = \begin{cases} tanh(\frac{q_i k_j^T}{\sqrt{d}}) & A_{ij} = 0 \\ \quad 0 & A_{ij} = 1 \end{cases}   
    \quad & , \quad w_{ij} & = \frac{e^{u_{ij}}}{\sum_{j=1}^{n} e^{u_{ij}}},
\end{alignat}
where \(j \in (1, \dots, n)\), \(n\) denotes the number of neighbors of node \(i\), \(A_{ij}\) is the adjacency matrix.

\subsection{Training}
For training, we adopt Soft Actor-Critic (SAC)~\cite{haarnoja2018soft}, an off-policy Actor-Critic method based on the maximum entropy principle, to optimize the policy network \(\pi_{\theta}(a_t \mid o_t)\) and critic network \(Q_{\phi}(o_t, a_t)\).

\noindent \textbf{Reward Function} To encourage effective exploration, the agent receives an immediate reward after each action $a_t$, consisting of three terms. The first term \( r_o = F_o(s_{t+1}) \) counts newly observed frontiers. The second term \( r_c = -F_c(s_t, s_{t+1}) \) penalizes the distance traveled. A terminal reward $r_f$ is given at the end of the episode only if the exploration is successfully completed. The total reward is thus:
\begin{equation}
r_t(o_t, a_t) = a \cdot r_o + b \cdot r_c + r_f,
\end{equation}
where \( a \) and \( b \) are scaling parameters (in practice \( a \) = 1/64, \( b \) = 1/50, \( r_f \) = 20).

\noindent \textbf{Soft Actor-Critic} The aim of SAC is to learn an optimal policy that maximizes the long-term reward while keeping its entropy as high as possible:
\begin{equation}
    \pi^* = \arg \max_{\pi} \mathbb{E}_{\pi} \left[ \sum_{t=1}^{T} \gamma^{t-1} \left( r_t + \alpha \mathcal{H} (\pi(\cdot | o_t)) \right) \right],
\end{equation}
where \( \pi^* \) is the optimal policy, \( T \) is the decision step, \( \gamma \) is the discount factor, \(r_t\) is the immediate reward, and \( \alpha \) is the temperature parameter that regulates the importance of the entropy term relative to the reward term.

\noindent In SAC, the soft state-value function is computed as:
\begin{equation}
    V(o_t) = \mathbb{E}_{a_t} \left[ Q_{\phi}(o_t, a_t) - \alpha \log(\pi_{\theta}(a_t | o_t)) \right].
\end{equation}

\noindent The critic loss is computed as:
\begin{equation}
    J(\phi) = \mathbb{E}_{o_t} \left[ \frac{1}{2} \left( Q_\phi(o_t, a_t) - \left( r_t + \gamma \mathbb{E}_{o_{t+1}} \left[ V(o_{t+1}) \right] \right) \right)^2 \right].
\end{equation}

\noindent The policy loss is computed as:
\begin{equation}
    J_{\pi}(\theta) = \mathbb{E}_{(o_t, a_t)} \left[ \alpha \log(\pi_\theta(a_t | o_t)) - Q_\phi(o_t, a_t) \right].
\end{equation}

\noindent The temperature parameter is automatically adjusted during training, and the temperature loss is computed as:
\begin{equation}
    J(\alpha) = \mathbb{E}_{a_t} \left[ -\alpha \left( \log \pi_{\theta}(a_t | o_t) + \overline{\mathcal{H}} \right) \right],
\end{equation}
where \(\overline{\mathcal{H}}\) represents the target entropy.

\noindent \textbf{Training Details} Our model is trained on simulated environments generated by a random dungeon generator~\cite{chen2020autonomous}. During training, the maximum episode length is set to 128 decision steps, with a discount factor $\gamma = 1$, batch size of 128, and an episode replay buffer size of 10{,}000. Training starts once the buffer contains more than 2{,}000 steps of data. The target entropy is defined as $0.05 \cdot \log(k)$. At the end of each episode, training is performed for 8 iterations using the Adam optimizer, where the learning rate is $5 \times 10^{-5}$ for both policy and critic networks, and $10^{-4}$ for temperature auto-tuning. The target critic network is updated every 64 training steps. All experiments are conducted on a desktop equipped with an Intel i5-14600KF CPU and an NVIDIA GeForce RTX 4070Ti Super GPU. We employ Ray~\cite{moritz2018ray}, a distributed framework to parallelize data collection by launching 16 agents simultaneously. Training converges after approximately 12{,}000 episodes, which takes about 36 hours.  

\subsection{Kalman Filter for Trajectory Smoothing}
In the training environment, the robot is abstracted as a node in the collision-free informative graph that can freely select its next waypoint (its k-nearest connected neighbors) from any direction in 360 degrees and transition to it instantaneously. However, testing in high-fidelity simulation environments reveals that this assumption may lead the planner to generate zig-zag or conflicting waypoints. While such behavior may be spatially optimal in the training environment, it fails to account for the motion constraints present in realistic settings. In both simulation and real-world environments, the robot cannot instantly reorient and transition; it must decelerate to a full stop, adjust its heading, and then accelerate toward the new target. Consequently, RL policy trained under this assumption are often not temporally optimal. To address this issue, we introduce a Kalman filter to smooth the output waypoints of the RL policy. Specifically, the historically selected waypoints are treated as observations, and the Kalman filter is applied to predict the probabilistic distribution of the robot position. The final next waypoint is selected by integrating the probabilistic outputs from both the RL policy and the Kalman filter.

\noindent \textbf{Kalman filter} is a recursive state estimator for linear dynamical systems with Gaussian noise. In a 2D scenario, we often track the position and velocity of a moving object in the \(x-y\) plane. In 2D Kalman filter, we define the 4-dimensional state vector at time step $k$ as: \(\mathbf{x}_k = \begin{bmatrix} x_k, y_k, v_{x,k}, v_{y,k} \end{bmatrix}^\top \in \mathbb{R}^4\), where \(x_k, y_k\) are the 2D positions, \(v_{x,k}, v_{y,k}\) are the 2D velocities.

\noindent The state evolves according to the linear dynamical system:
\begin{equation}
    \mathbf{x}_k = \mathbf{F} \mathbf{x}_{k-1} + \mathbf{w}_{k-1}, \quad \mathbf{w}_{k-1} \sim \mathcal{N}(0, \mathbf{Q}),
\end{equation}
where \(\mathbf{F}\) is the state transition matrix, \(\mathbf{w}_{k-1}\) is the process noise, modeling acceleration and model uncertainty, \(\mathbf{Q}\) is the process noise covariance matrix.

\noindent We assume that only the position $(x_k, y_k)$ is observable:
\begin{equation}
    \mathbf{z}_k = \mathbf{H} \mathbf{x}_k + \mathbf{v}_k, \quad \mathbf{v}_k \sim \mathcal{N}(0, \mathbf{R}),
\end{equation}
where \(\mathbf{H}\) is the observation matrix, \(\mathbf{z}_k\) is the observation vector, \(\mathbf{v}_k\) is the measurement noise, \(\mathbf{R}\) is the measurement noise covariance matrix.

\noindent The Kalman filter prediction process:
\begin{align}
\text{Predicted state:} \quad  \hat{\mathbf{x}}_k^- &= \mathbf{F} \hat{\mathbf{x}}_{k-1} \notag \\
\text{Predicted covariance:} \quad  \mathbf{P}_k^- &= \mathbf{F} \mathbf{P}_{k-1} \mathbf{F}^\top + \mathbf{Q}.
\end{align}

\noindent The Kalman filter update process:
\begin{align}
\text{Kalman gain:} \quad  \mathbf{K}_k &= \mathbf{P}_k^- \mathbf{H}^\top (\mathbf{H} \mathbf{P}_k^- \mathbf{H}^\top + \mathbf{R})^{-1} \notag \\
\text{Updated state:} \quad  \hat{\mathbf{x}}_k &= \hat{\mathbf{x}}_k^- + \mathbf{K}_k (\mathbf{z}_k - \mathbf{H} \hat{\mathbf{x}}_k^-) \\
\text{Updated covariance:} \quad  \mathbf{P}_k &= (\mathbf{I} - \mathbf{K}_k \mathbf{H}) \mathbf{P}_k^- \notag.
\end{align}

\noindent At each time step, the posterior belief over the state is a multivariate Gaussian distribution:
\begin{equation}
    \mathbf{x}_k \sim \mathcal{N}(\hat{\mathbf{x}}_k, \mathbf{P}_k).
\end{equation}

\noindent The marginal distribution over position is:
\begin{equation}
    \mathbf{x}_k^{\text{pos}} =
    \begin{bmatrix}
    x_k \\
    y_k
    \end{bmatrix}
    \sim \mathcal{N}(\hat{\mathbf{x}}_k^{[:2]}, \mathbf{P}_k^{[:2,:2]}).
\end{equation}

\noindent To compute the probability density of a given position $\mathbf{p} = [x, y]^\top$ under the Gaussian distribution:
\begin{equation}
    f(\mathbf{p}) = \frac{1}{2\pi |\mathbf{P}|^{1/2}} 
    \exp\left( -\frac{1}{2} (\mathbf{p} - \boldsymbol{\mu})^\top \mathbf{P}^{-1} (\mathbf{p} - \boldsymbol{\mu}) \right),
\end{equation}
where \(\boldsymbol{\mu} = \hat{\mathbf{x}}_k^{[:2]}\) is the estimated position mean, \(\mathbf{P} = \mathbf{P}_k^{[:2,:2]}\) is the position covariance matrix.

\noindent The final output policy \(\pi'(a_t | o_t)\) can be calculated as:
\begin{equation}
    \pi'(a_t | o_t) = \pi(a_t | o_t) + \frac{f(s_{t+1} | s_t,a_t)}{\sum_{a_t \in A} f(s_{t+1} | s_t,a_t)}.
\end{equation}

\begin{table*}[t]
\caption{\textbf{Comparisons with baseline planners in small-scale training environments (Average travel distance of 200 scenarios)}}
\vspace{-0.5cm}
\label{table_1}
\begin{center}
\begin{tabular}{c|c|c|c|c|c|c}
    \toprule 
    & Nearest & Utility & NBVP & TARE Local &  ARiADNE & \textbf{Ours}\\
    \midrule 
    Distance ($m$) & 490 ($\pm$ 180) & 442 ($\pm$ 163) & 504 ($\pm$ 201) & 382 ($\pm$ 156) & 402 ($\pm$ 201) & \textbf{372 ($\pm$ 166)} \\
    \bottomrule 
\end{tabular}
\end{center}
\vspace{-0.3cm}
\end{table*}

\begin{figure*}[t]
    \centering
    \setlength{\tabcolsep}{5pt} 
    \begin{tabular}{ccc}
        \includegraphics[height=5.5cm]{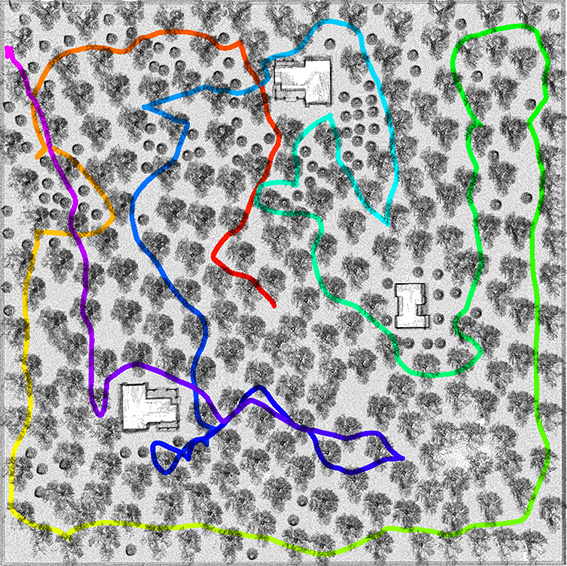} &
        \includegraphics[height=5.5cm]{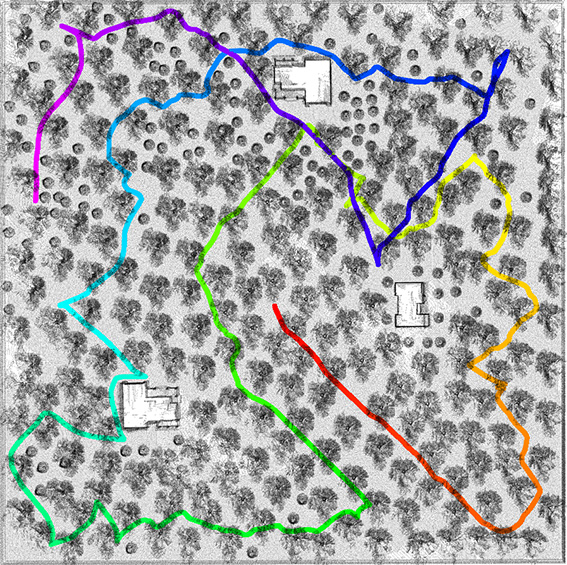} &
        \includegraphics[height=5.5cm]{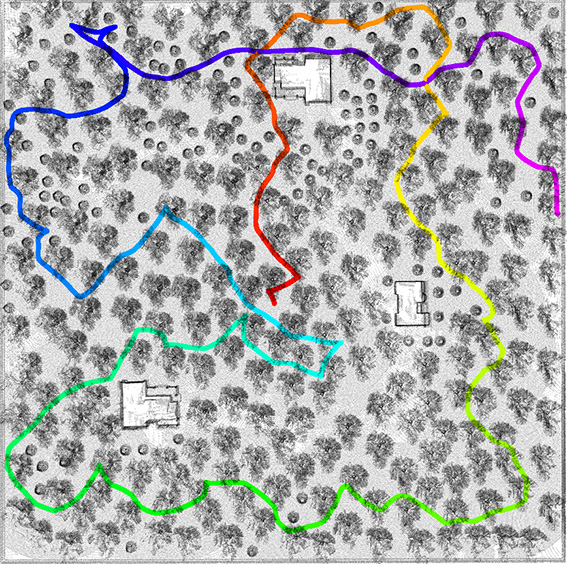}\\
        {\footnotesize (a) TARE, \(\SI{1418}{m}, \SI{741}{s}, \SI{1.91}{m/s}\)} & {\footnotesize (b) ARiADNE, \(\SI{1137}{m}, \SI{668}{s}, \SI{1.70}{m/s}\)} & {\footnotesize (c) GRATE, \(\SI{1026}{m}, \SI{554}{s}, \SI{1.85}{m/s}\)}\\
    \end{tabular}
    \caption{\textbf{Exploration efficiency comparisons in large-scale forest Gazebo simulation environment ($\SI{150}{m} \times \SI{150}{m}$) provided by~\cite{cao2022autonomous}.}}
    \label{fig_3}
    \vspace{-0.55cm}
\end{figure*}

\section{Experiments}
\subsection{Training Environment}
We first evaluated our trained RL policy on a test set consisting of 200 previously unseen simulated 2D dungeon environments. Since only the robot's position transitions are considered during training, no Kalman Filter is applied for trajectory smoothing. We compared the travel distance of GRATE to complete the exploration with several conventional and learning-based exploration planners: (1) Nearest~\cite{yamauchi1997frontier}: the robot greedily moves to the nearest frontier, (2) Utility~\cite{gonzalez2002navigation}: the robot selects frontiers by balancing exploration cost and expected utility, (3) NBVP~\cite{bircher2016NBVP}: a sampling-based planner relying on rapidly-exploring random trees, (4) TARE Local~\cite{cao2021tare}: the local planner of TARE (excluding the global planner, as its local horizon suffices for our test scenarios), (5) ARiADNE~\cite{cao2023ariadne}: an attention-based DRL planner. As shown in Table \ref{table_1}, GRATE outperforms all baseline planners in terms of travel distance. In particular, GRATE achieves a 7.5\% improvement over ARiADNE, highlighting its stronger reasoning capability across the entire map.

\begin{figure*}[t]
    \centering
    \setlength{\tabcolsep}{3pt} 
    \begin{tabular}{ccc}
        \includegraphics[height=5cm]{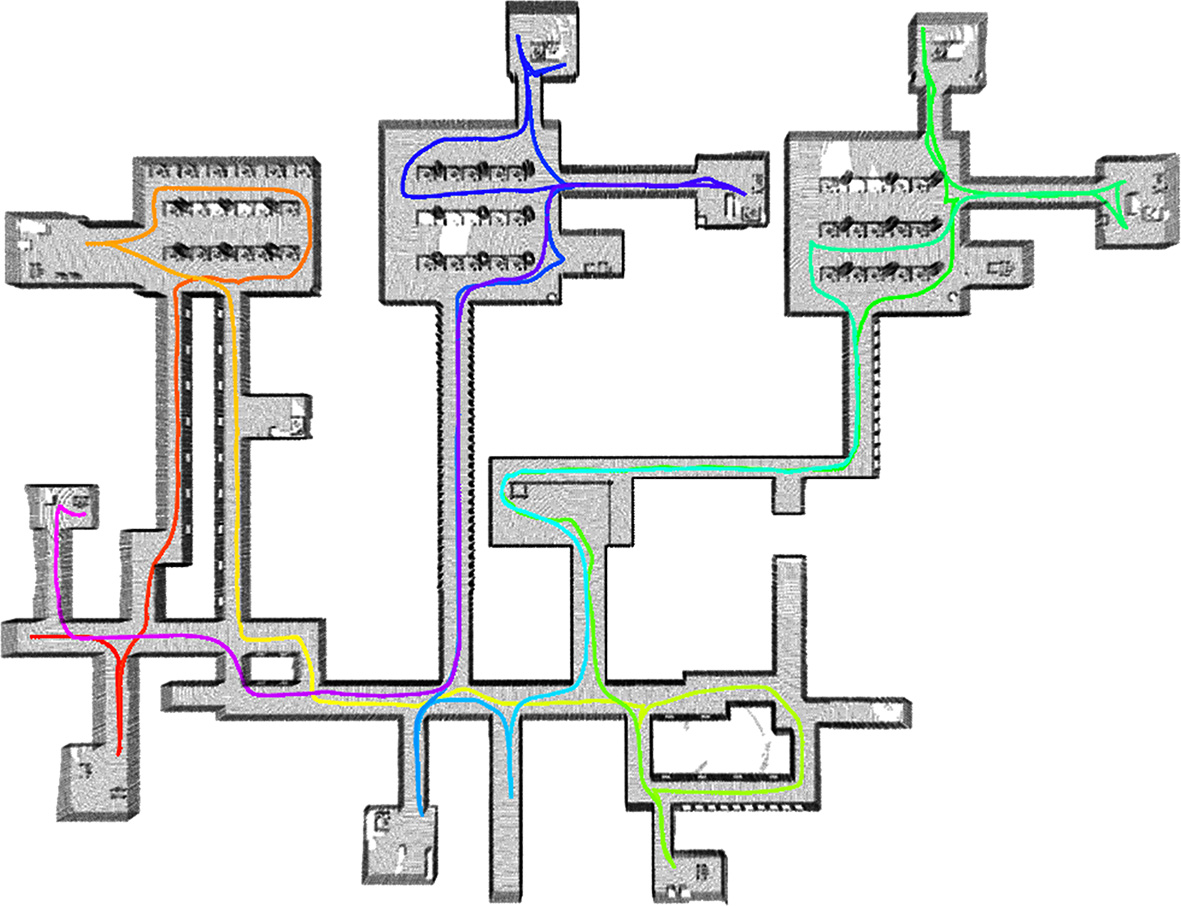} \hspace{-0.8cm} &
        \includegraphics[height=5cm]{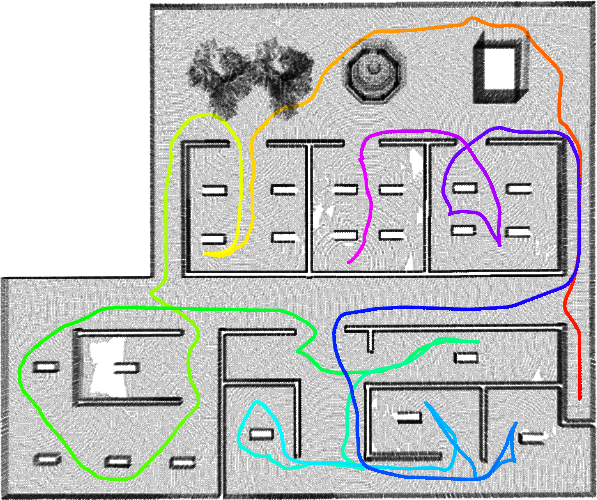} &
        \includegraphics[height=5cm]{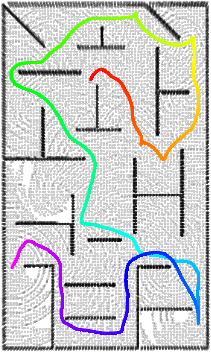}\\
        {\footnotesize (a) large-scale, GRATE, \(\SI{985}{m}, \SI{544}{s}, \SI{1.81}{m/s}\)} & {\footnotesize (b) medium-scale, GRATE, \(\SI{554}{m}, \SI{307}{s}, \SI{1.80}{m/s}\)} & {\footnotesize (c) small-scale, GRATE, \(\SI{127}{m}, \SI{100}{s}, \SI{1.27}{m/s}\)}\\
    \end{tabular}
    \caption{\textbf{Visualization Results of GRATE in indoor Gazebo simulation environments.} Left: large-scale indoor environment ($\SI{130}{m} \times \SI{100}{m}$); Middle: medium-scale indoor environment ($\SI{85}{m} \times \SI{64}{m}$); Right: small-scale indoor environment ($\SI{35}{m} \times \SI{25}{m}$).}
    \label{fig_4}
    \vspace{-0.5cm}
\end{figure*}

\begin{figure}[t]
    \vspace{0.2cm}
    \centering
    \setlength{\tabcolsep}{2.8pt} 
    \begin{tabular}{ccc}
        \includegraphics[height=2.3cm]{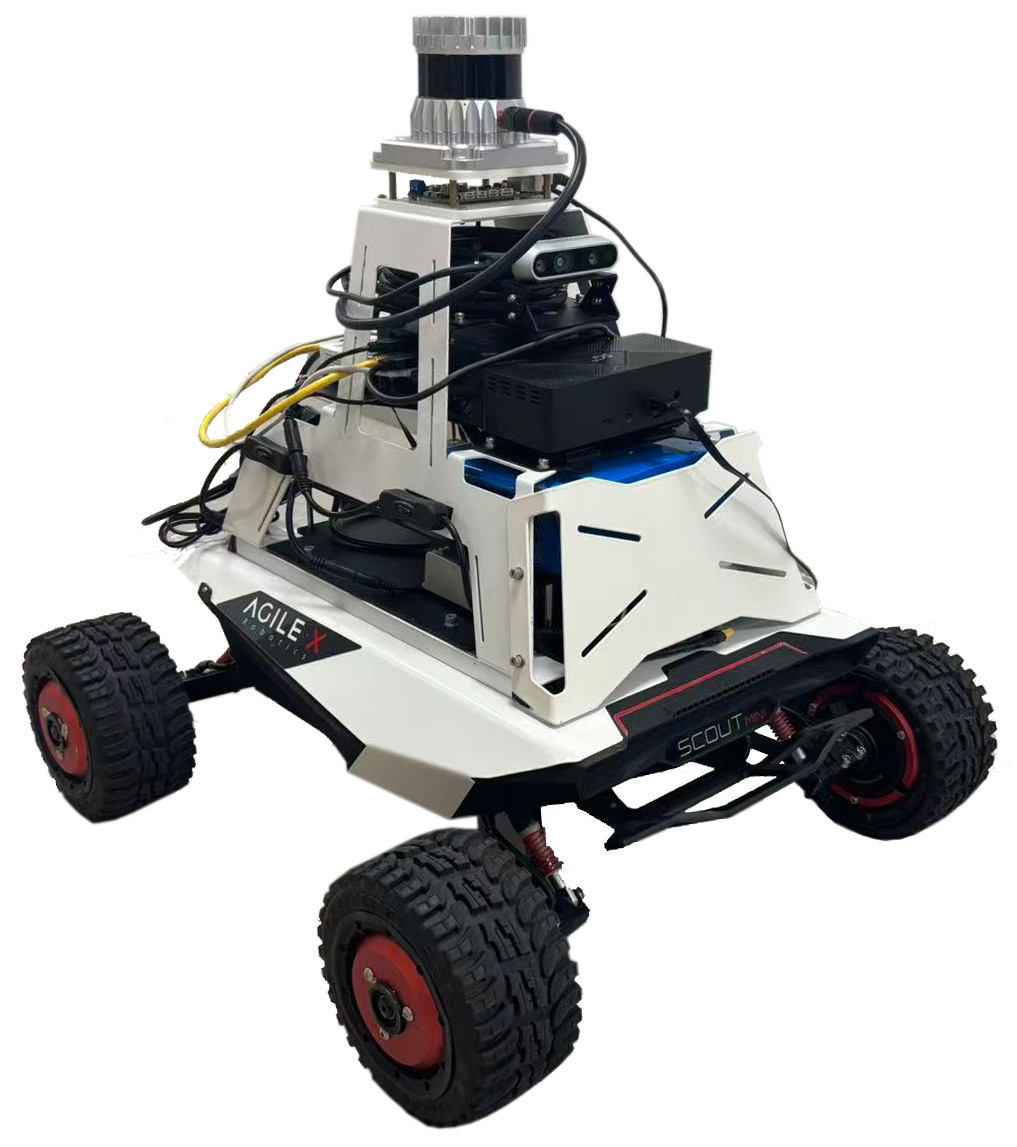} &
        \includegraphics[height=2.3cm]{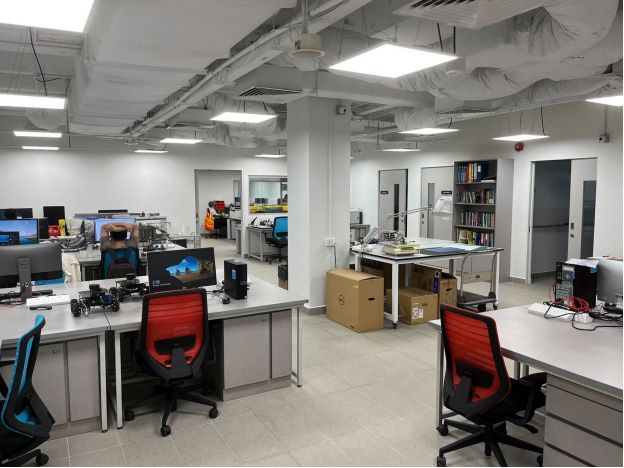} &
        \includegraphics[height=2.3cm]{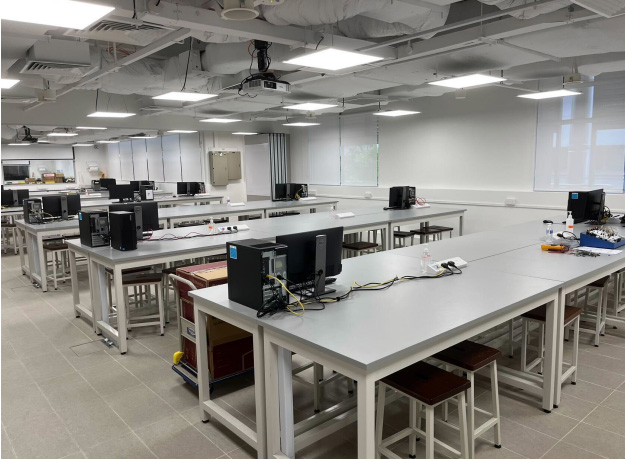}\\
        {\footnotesize (a) mobile robot} & {\footnotesize (b) indoor scene-1} & {\footnotesize (c) indoor scene-2} \vspace{0.2cm} \\
    \end{tabular}
    \begin{tabular}{c}
        \includegraphics[width=1\linewidth]{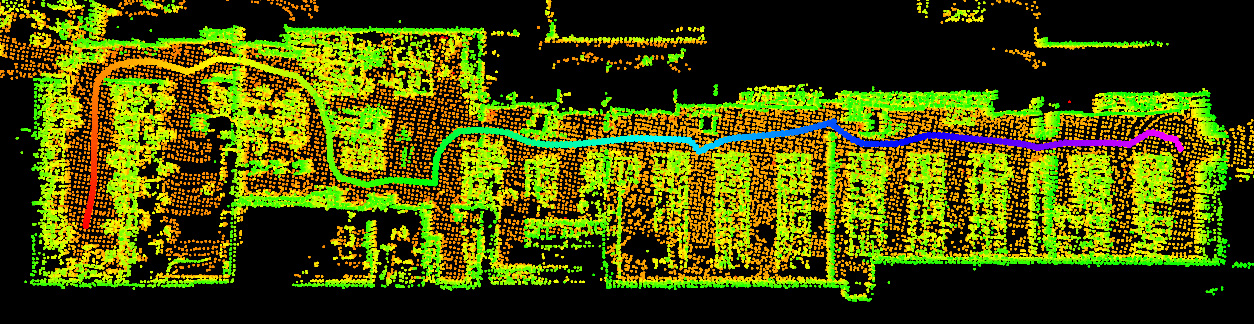}\\
        {\footnotesize (d) final point cloud map}\\
    \end{tabular}
    \caption{Real world validation in a $\SI{60}{m} \times \SI{10}{m}$ indoor laboratory.}
    \label{fig_5}
    \vspace{-0.7cm}
\end{figure}

\begin{table}[t]
\vspace{0.2cm}
\caption{\textbf{Comparisons with baseline planners in Gazebo simulation environments (5 runs each).}}
\vspace{-0.5cm}
\label{table_2}
\begin{center}
\begin{tabular}{c|c|c|c|c}
    \toprule
    Scene & Method & \makecell[c]{Distance \\ (m)} & Time (s) & \makecell[c]{Velocity \\(m/s)} \\
    \midrule
    \multirow{3}{*}{\makecell[c]{Large-scale \\ forest \\ ($\SI{150}{m} \times \SI{150}{m}$)}}
      & TARE & 1368 & 720 & \textbf{1.90} \\
      & ARiADNE  & 1158  & 675  & 1.72  \\
      & GRATE & \textbf{1074} & \textbf{577} & 1.86 \\
    \midrule
    \multirow{3}{*}{\makecell[c]{Large-scale \\ indoor \\ ($\SI{130}{m} \times \SI{100}{m}$)}}
      & TARE & 1178 & 644 & \textbf{1.83} \\
      & ARiADNE & 1042  & 602  & 1.73  \\
      & GRATE & \textbf{980} & \textbf{542} & 1.81 \\
    \midrule
    \multirow{3}{*}{\makecell[c]{Medium-scale \\ indoor \\ ($\SI{85}{m} \times \SI{64}{m}$)}}
      & FAEL & 592 & 365 & 1.62 \\
      & ARiADNE & \textbf{540} & 328 & 1.69 \\
      & GRATE & 548 & \textbf{303} & \textbf{1.81} \\
    \midrule
    \multirow{3}{*}{\makecell[c]{Small-scale \\ indoor \\ ($\SI{35}{m} \times \SI{25}{m}$)}}
      & HPHS & 136 & 122 & 1.11 \\
      & ARiADNE & 129 & 105 & 1.23 \\
      & GRATE & \textbf{125} & \textbf{96} & \textbf{1.30} \\
    \bottomrule
\end{tabular}
\end{center}
\vspace{-0.6cm}
\end{table}

\subsection{Gazebo Simulation}
We then tested GRATE in four high-fidelity 3D Gazebo simulation benchmarks of different scales: a large-scale forest environment ($\SI{150}{m} \times \SI{150}{m}$) and a large-scale indoor environment ($\SI{130}{m} \times \SI{100}{m}$) provided by~\cite{cao2022autonomous} (also the author of TARE~\cite{cao2021tare, cao2023representation}), a medium-scale indoor environment ($\SI{85}{m} \times \SI{64}{m}$) provided by FAEL~\cite{huang2023fael}, and a small-scale indoor environment ($\SI{35}{m} \times \SI{25}{m}$) provided by HPHS~\cite{long2024hphs}. Unlike the abstract 2D dungeon environments employed during training, which adopt simplified assumptions about sensor perception and robot kinematics, the Gazebo simulation benchmarks provide high-fidelity testing scenarios. They integrate a physically realistic sensor model (Velodyne 16-line 3D LiDAR) together with practical robot motion constraints, represented by a four-wheel differential-drive platform capable of reaching a maximum velocity of \SI{2}{m/s}. Current robot's belief (fused point cloud data collected from the sensor with Simultaneous Localization and Mapping (SLAM) provided by~\cite{cao2022autonomous}) is classified into free and occupied areas using Octomap~\cite{hornung2013octomap}. The resulting map is then projected to a 2D occupancy grid and further extracted into a sparse informative graph~\cite{cao2024deep}, enabling the extension of our trained RL policy from small-scale to large-scale environments. Finally, the next waypoint (computed by integrating the probability outputs of the trained RL policy and the Kalman filter) is tracked by the robot using the local motion planner described in~\cite{cao2022autonomous}.

We compared GRATE with: (1) TARE~\cite{cao2021tare, cao2023representation}, (2) FAEL~\cite{huang2023fael}, (3) HPHS~\cite{long2024hphs}, (4) ARiADNE~\cite{cao2023ariadne, cao2024deep}. We evaluate the performance of all planners from multiple metrics: (1) travel distance, (2) exploration time, and (3) average velocity. Results are shown in Fig.\ref{fig_3}, Fig.\ref{fig_4} and Table.\ref{table_2}. Specifically, in large-scale forest environment, GRATE outperforms TARE by 21.5\% in travel distance and 19.9\% in exploration time, and ARiADNE by 7.3\% and 14.5\%, respectively. In large-scale indoor environments, it achieves 16.8\% and 15.8\% improvements over TARE, and 6.0\% and 10.0\% over ARiADNE. In medium-scale indoor environments, GRATE reduces exploration time by 17.0\% compared to FAEL and by 7.6\% compared to ARiADNE, while in small-scale environments, it achieves improvements of 21.3\% over HPHS and 8.6\% over ARiADNE. In terms of velocity, GRATE achieves on par performance with TARE and consistently better than ARiADNE, FAEL, and HPHS. Overall, GRATE achieves substantial improvements in temporal efficiency and performs comparable or superior spatial efficiency relative to both conventional and learning-based SOTA methods.

It is important to note that TARE explicitly incorporates curvature constraints into its local planner, which smooths the generated path and guarantees kinodynamic feasibility for path following. Consequently, TARE’s velocity performance can be regarded as near-optimal. That GRATE achieves comparable average velocity to TARE demonstrates the effectiveness of the Kalman filter in GRATE, which contributes to waypoint smoothing and enhanced temporal efficiency. Moreover, GRATE outperforms ARiADNE in terms of exploration time and average velocity, further demonstrating the effectiveness of Kalman filter in GRATE. GRATE also achieves a significant improvement than ARiADNE in terms of travel distance, which is consistent with our evaluation results in the 2D dungeon training environments. This improvement stems from the fact that ARiADNE incorporate a structurally masked multi-head self-attention mechanism guided by graph adjacency, restricting each node’s receptive field to its neighbors and requiring multiple layers for global propagation. This design performs well on small-scale graphs, but as the graph grows, stacking additional layers becomes necessary to expand the receptive field. However, over-squashing limits each node’s fixed-size feature from adequately capturing global structural information, leaving distant nodes underrepresented. Meanwhile, relying solely on the global attention mechanism in Transformers overlooks the structural information encoded by graph adjacency. The Graph Transformer framework in GRATE address these limitations by integrating local message-passing GNNs with Transformers, effectively leveraging GNNs to encode local structural information while employing Transformers to model long-term dependencies.

\vspace{-0.1cm}
\subsection{Real World Validation}

We further deployed GRATE on an AgileX Scout mini mobile robot equipped with a Ouster 32-line 3D LiDAR. The experiment was conducted in a $\SI{60}{m} \times \SI{10}{m}$ indoor laboratory with cluttered furniture (Fig.\ref{fig_5}). Similar to the simulation setup, we adopted the SLAM algorithm provided by~\cite{cao2022autonomous} to obtain both LiDAR odometry and mapping for real-world validation. The generated waypoints were tracked using the local motion planner provided by~\cite{cao2022autonomous}, with the robot’s max speed set to \SI{0.75}{m/s}. For our DRL planner, we configured the effective sensor range to \SI{20}{m}, the Octomap resolution to \SI{0.2}{m}, and the node resolution of the extracted collision-free informative graph to \SI{0.8}{m}. In this experiment, GRATE successfully explored the environment, achieving a travel distance of \SI{67}{m}, an exploration time of \SI{167}{s}, thereby demonstrating the robustness of our DRL planner in real-world scenarios.

\section{Conclusion}
\addtolength{\textheight}{0cm}

In this work, we introduce GRATE, a DRL-based approach for autonomous robot exploration. GRATE relies on a novel Graph Transformer-based encoder to jointly capture local and global dependencies, thereby enhancing the model’s capacity to reason over both explored and unexplored regions. We further develop a Kalman filter to smooth the waypoint output of the RL policy and enhance motion continuity by reasoning about the robot's kinodynamic constraints, which achieves notable improvements in temporal efficiency. In our experimental results, GRATE outperforms SOTA conventional and learning-based approaches, demonstrating substantial improvements in exploration efficiency across diverse simulation benchmarks as well as real-world scenarios. Future work will target the deployment of our planner in multi-robot settings, with a particular emphasis on promoting collaborative efficiency while reasoning about motion constraints for multi-robot collision avoidance.

\bibliographystyle{unsrt}
\bibliography{Reference}

\begin{thebibliography}{10}

\bibitem{yamauchi1997frontier}
Brian Yamauchi.
\newblock A frontier-based approach for autonomous exploration.
\newblock In {\em Proceedings 1997 IEEE International Symposium on
  Computational Intelligence in Robotics and Automation CIRA'97.'Towards New
  Computational Principles for Robotics and Automation'}, pages 146--151. IEEE,
  1997.

\bibitem{gonzalez2002navigation}
H{\'e}ctor~H Gonz{\'a}lez-Banos and Jean-Claude Latombe.
\newblock Navigation strategies for exploring indoor environments.
\newblock {\em The International Journal of Robotics Research},
  21(10-11):829--848, 2002.

\bibitem{osswald2016speeding}
Stefan O{\ss}wald, Maren Bennewitz, Wolfram Burgard, and Cyrill Stachniss.
\newblock Speeding-up robot exploration by exploiting background information.
\newblock {\em IEEE Robotics and Automation Letters}, 1(2):716--723, 2016.

\bibitem{cieslewski2017rapid}
Titus Cieslewski, Elia Kaufmann, and Davide Scaramuzza.
\newblock Rapid exploration with multi-rotors: A frontier selection method for
  high speed flight.
\newblock In {\em 2017 IEEE/RSJ International Conference on Intelligent Robots
  and Systems (IROS)}, pages 2135--2142. IEEE, 2017.

\bibitem{cao2023ariadne}
Yuhong Cao, Tianxiang Hou, Yizhuo Wang, Xian Yi, and Guillaume Sartoretti.
\newblock Ariadne: A reinforcement learning approach using attention-based deep
  networks for exploration.
\newblock In {\em 2023 IEEE International Conference on Robotics and Automation
  (ICRA)}, pages 10219--10225. IEEE, 2023.

\bibitem{bircher2016NBVP}
Andreas Bircher, Mina Kamel, Kostas Alexis, Helen Oleynikova, and Roland
  Siegwart.
\newblock Receding horizon" next-best-view" planner for 3d exploration.
\newblock In {\em 2016 IEEE international conference on robotics and automation
  (ICRA)}, pages 1462--1468. IEEE, 2016.

\bibitem{cao2021tare}
Chao Cao, Hongbiao Zhu, Howie Choset, and Ji~Zhang.
\newblock Tare: A hierarchical framework for efficiently exploring complex 3d
  environments.
\newblock In {\em Robotics: Science and Systems}, volume~5, page~2, 2021.

\bibitem{cao2023representation}
Chao Cao, Hongbiao Zhu, Zhongqiang Ren, Howie Choset, and Ji~Zhang.
\newblock Representation granularity enables time-efficient autonomous
  exploration in large, complex worlds.
\newblock {\em Science Robotics}, 8(80):eadf0970, 2023.

\bibitem{huang2023fael}
Junlong Huang, Boyu Zhou, Zhengping Fan, Yilin Zhu, Yingrui Jie, Longwei Li,
  and Hui Cheng.
\newblock Fael: Fast autonomous exploration for large-scale environments with a
  mobile robot.
\newblock {\em IEEE robotics and automation letters}, 8(3):1667--1674, 2023.

\bibitem{long2024hphs}
Shijun Long, Ying Li, Chenming Wu, Bin Xu, and Wei Fan.
\newblock Hphs: hierarchical planning based on hybrid frontier sampling for
  unknown environments exploration.
\newblock In {\em 2024 IEEE/RSJ International Conference on Intelligent Robots
  and Systems (IROS)}, pages 12056--12063. IEEE, 2024.

\bibitem{cao2024deep}
Yuhong Cao, Rui Zhao, Yizhuo Wang, Bairan Xiang, and Guillaume Sartoretti.
\newblock Deep reinforcement learning-based large-scale robot exploration.
\newblock {\em IEEE Robotics and Automation Letters}, 2024.

\bibitem{cao2022autonomous}
Chao Cao, Hongbiao Zhu, Fan Yang, Yukun Xia, Howie Choset, Jean Oh, and
  Ji~Zhang.
\newblock Autonomous exploration development environment and the planning
  algorithms.
\newblock In {\em 2022 International Conference on Robotics and Automation
  (ICRA)}, pages 8921--8928. IEEE, 2022.

\bibitem{liang2024hdplanner}
Jingsong Liang, Yuhong Cao, Yixiao Ma, Hanqi Zhao, and Guillaume Sartoretti.
\newblock Hdplanner: Advancing autonomous deployments in unknown environments
  through hierarchical decision networks.
\newblock {\em IEEE Robotics and Automation Letters}, 2024.

\bibitem{chen2020autonomous}
Fanfei Chen, John~D Martin, Yewei Huang, Jinkun Wang, and Brendan Englot.
\newblock Autonomous exploration under uncertainty via deep reinforcement
  learning on graphs.
\newblock In {\em 2020 IEEE/RSJ International Conference on Intelligent Robots
  and Systems (IROS)}, pages 6140--6147. IEEE, 2020.

\bibitem{velickovic2018gan}
Petar Veličković, Guillem Cucurull, Arantxa Casanova, Adriana Romero, Pietro
  Liò, and Yoshua Bengio.
\newblock Graph attention networks, 2018.

\bibitem{alon2021bottleneck}
Uri Alon and Eran Yahav.
\newblock On the bottleneck of graph neural networks and its practical
  implications.
\newblock In {\em International Conference on Learning Representations}, 2021.

\bibitem{topping2022understanding}
Jake Topping, Francesco Di~Giovanni, Benjamin~Paul Chamberlain, Xiaowen Dong,
  and Michael~M Bronstein.
\newblock Understanding over-squashing and bottlenecks on graphs via curvature.
\newblock In {\em International Conference on Learning Representations}, 2022.

\bibitem{vaswani2017attention}
Ashish Vaswani, Noam Shazeer, Niki Parmar, Jakob Uszkoreit, Llion Jones,
  Aidan~N Gomez, {\L}ukasz Kaiser, and Illia Polosukhin.
\newblock Attention is all you need.
\newblock {\em Advances in neural information processing systems}, 30, 2017.

\bibitem{wu2021representing}
Zhanghao Wu, Paras Jain, Matthew Wright, Azalia Mirhoseini, Joseph~E Gonzalez,
  and Ion Stoica.
\newblock Representing long-range context for graph neural networks with global
  attention.
\newblock {\em Advances in neural information processing systems},
  34:13266--13279, 2021.

\bibitem{chen2022structure}
Dexiong Chen, Leslie O’Bray, and Karsten Borgwardt.
\newblock Structure-aware transformer for graph representation learning.
\newblock In {\em International conference on machine learning}, pages
  3469--3489. PMLR, 2022.

\bibitem{rampavsek2022recipe}
Ladislav Ramp{\'a}{\v{s}}ek, Michael Galkin, Vijay~Prakash Dwivedi, Anh~Tuan
  Luu, Guy Wolf, and Dominique Beaini.
\newblock Recipe for a general, powerful, scalable graph transformer.
\newblock {\em Advances in Neural Information Processing Systems},
  35:14501--14515, 2022.

\bibitem{wu2023sgformer}
Qitian Wu, Wentao Zhao, Chenxiao Yang, Hengrui Zhang, Fan Nie, Haitian Jiang,
  Yatao Bian, and Junchi Yan.
\newblock Sgformer: Simplifying and empowering transformers for large-graph
  representations.
\newblock {\em Advances in Neural Information Processing Systems},
  36:64753--64773, 2023.

\bibitem{dwivedi2023benchmarking}
Vijay~Prakash Dwivedi, Chaitanya~K Joshi, Anh~Tuan Luu, Thomas Laurent, Yoshua
  Bengio, and Xavier Bresson.
\newblock Benchmarking graph neural networks.
\newblock {\em Journal of Machine Learning Research}, 24(43):1--48, 2023.

\bibitem{zhu2018deep}
Delong Zhu, Tingguang Li, Danny Ho, Chaoqun Wang, and Max Q-H Meng.
\newblock Deep reinforcement learning supervised autonomous exploration in
  office environments.
\newblock In {\em 2018 IEEE international conference on robotics and automation
  (ICRA)}, pages 7548--7555. IEEE, 2018.

\bibitem{chen2019self}
Fanfei Chen, Shi Bai, Tixiao Shan, and Brendan Englot.
\newblock Self-learning exploration and mapping for mobile robots via deep
  reinforcement learning.
\newblock In {\em Aiaa scitech 2019 forum}, page 0396, 2019.

\bibitem{wang2025cogniplan}
Yizhuo Wang, Haodong He, Jingsong Liang, Yuhong Cao, Ritabrata Chakraborty, and
  Guillaume Sartoretti.
\newblock Cogniplan: Uncertainty-guided path planning with conditional
  generative layout prediction.
\newblock {\em arXiv preprint arXiv:2508.03027}, 2025.

\bibitem{harutyunyan2025mapexrl}
Narek Harutyunyan, Brady Moon, Seungchan Kim, Cherie Ho, Adam Hung, and
  Sebastian Scherer.
\newblock Mapexrl: Human-inspired indoor exploration with predicted environment
  context and reinforcement learning.
\newblock {\em arXiv preprint arXiv:2503.01548}, 2025.

\bibitem{yuan2025survey}
Chaohao Yuan, Kangfei Zhao, Ercan~Engin Kuruoglu, Liang Wang, Tingyang Xu,
  Wenbing Huang, Deli Zhao, Hong Cheng, and Yu~Rong.
\newblock A survey of graph transformers: Architectures, theories and
  applications, 2025.

\bibitem{ying2021transformers}
Chengxuan Ying, Tianle Cai, Shengjie Luo, Shuxin Zheng, Guolin Ke, Di~He,
  Yanming Shen, and Tie-Yan Liu.
\newblock Do transformers really perform badly for graph representation?
\newblock {\em Advances in neural information processing systems},
  34:28877--28888, 2021.

\bibitem{dwivedi2020generalization}
Vijay~Prakash Dwivedi and Xavier Bresson.
\newblock A generalization of transformer networks to graphs.
\newblock {\em arXiv preprint arXiv:2012.09699}, 2020.

\bibitem{vinyals2015pointer}
Oriol Vinyals, Meire Fortunato, and Navdeep Jaitly.
\newblock Pointer networks.
\newblock {\em Advances in neural information processing systems}, 28, 2015.

\bibitem{hornung2013octomap}
Armin Hornung, Kai~M Wurm, Maren Bennewitz, Cyrill Stachniss, and Wolfram
  Burgard.
\newblock Octomap: An efficient probabilistic 3d mapping framework based on
  octrees.
\newblock {\em Autonomous robots}, 34(3):189--206, 2013.

\bibitem{xu2018powerful}
Keyulu Xu, Weihua Hu, Jure Leskovec, and Stefanie Jegelka.
\newblock How powerful are graph neural networks?
\newblock {\em arXiv preprint arXiv:1810.00826}, 2018.

\bibitem{bresson2017residual}
Xavier Bresson and Thomas Laurent.
\newblock Residual gated graph convnets.
\newblock {\em arXiv preprint arXiv:1711.07553}, 2017.

\bibitem{hochreiter1997long}
Sepp Hochreiter and J{\"u}rgen Schmidhuber.
\newblock Long short-term memory.
\newblock {\em Neural computation}, 9(8):1735--1780, 1997.

\bibitem{haarnoja2018soft}
Tuomas Haarnoja, Aurick Zhou, Pieter Abbeel, and Sergey Levine.
\newblock Soft actor-critic: Off-policy maximum entropy deep reinforcement
  learning with a stochastic actor.
\newblock In {\em International conference on machine learning}, pages
  1861--1870. Pmlr, 2018.

\bibitem{moritz2018ray}
Philipp Moritz, Robert Nishihara, Stephanie Wang, Alexey Tumanov, Richard Liaw,
  Eric Liang, Melih Elibol, Zongheng Yang, William Paul, Michael~I Jordan,
  et~al.
\newblock Ray: A distributed framework for emerging $\{$AI$\}$ applications.
\newblock In {\em 13th USENIX symposium on operating systems design and
  implementation (OSDI 18)}, pages 561--577, 2018.

\end{thebibliography}

\end{document}